\documentclass[11pt,a4paper]{article}
\usepackage[hyphens]{url}
\usepackage[hyperref]{emnlp2020}
\usepackage[hyphenbreaks]{breakurl}
\usepackage{cleveref}

\crefformat{section}{\S#2#1#3} %
\crefformat{subsection}{\S#2#1#3}
\crefformat{subsubsection}{\S#2#1#3}
\usepackage{times}
\usepackage{latexsym}

\usepackage{microtype}

\usepackage{todonotes}
\usepackage{xspace}

\newcommand{\fracas}{\textit{FraCas}\xspace}

\usepackage{comment}
\usepackage{booktabs}
\usepackage{diagbox}
\usepackage{adjustbox}
\usepackage{dcolumn}
\newcolumntype{H}{>{\setbox0=\hbox\bgroup}c<{\egroup}@{}}
\usepackage{multirow}
\usepackage{amssymb}

\aclfinalcopy %
\def\aclpaperid{23} %

\title{A Survey on Recognizing Textual Entailment as an NLP Evaluation}

\author{Adam Poliak \\
  Barnard College, Data Science Institute, Columbia University \\
  3009 Broadway, New York, NY 10027 \\
  \texttt{apoliak@barnard.ed} \\}

\date{}

\begin{document}
\maketitle
\begin{abstract}
Recognizing Textual Entailment (RTE) was proposed as a unified evaluation framework to compare semantic understanding of different NLP systems. In this survey paper, we provide an overview of different approaches for evaluating and understanding the reasoning capabilities of NLP systems. We then focus our discussion on RTE by highlighting prominent RTE datasets as well as advances in RTE dataset that focus on specific linguistic phenomena that can be used %
to evaluate NLP systems on a fine-grained level. 
We conclude by arguing that when evaluating NLP systems, the community should utilize newly introduced RTE datasets that focus on specific linguistic phenomena.
\end{abstract}

\section{Introduction}
As NLP technologies are more widely adopted, how to evaluate NLP systems and how to determine whether one model understands language or generates text better than another  %
is an increasingly important question.
Recognizing Textual Entailment~\cite[RTE][]{cooper1996using,dagan2006pascal}, the task of determining whether the meaning of one sentence can likely be inferred from another was introduced to answer this question.

We begin this survey by discussing different approaches over the past thirty years for evaluating and comparing NLP systems. %
Next, we will discuss how  
RTE was introduced as a specific answer to this
broad question of how to best evaluate NLP systems. 
This will include a broad discussion
of efforts in the past three decades to build RTE datasets
and use RTE to evaluate NLP models. We will then highlight recent RTE datasets that focus on specific semantic phenomena and conclude by arguing that they should be utilized for evaluating the reasoning capabilities of downstream NLP systems. 

\subsection*{Natural Language Inference or Recognizing Textual Entailment?}
The terms Natural Language Inference (NLI) and RTE are often used interchangeably. Many papers 
begin by explicitly mentioning that these terms are synonymous~\cite{liu2016learning,gong2018natural,NIPS2018_8163}.\footnote{In fact, variants of the phrase ``natural language inference, also known as recognizing textual entailment'' appear in many papers~\cite[][\textit{i.a.}]{chen-etal-2017-recurrent,williams2017broad,naik-EtAl:2018:C18-1,chen-etal-2018-neural,tay-etal-2018-compare}.}
The broad phrase ``natural language inference'' is more appropriate for a class of problems that require making inferences from natural language. Tasks like sentiment analysis, event factuality, or even question-answering can be viewed as forms of natural language inference without having to convert them into the sentence pair classification format used in RTE. %
Earlier works used the term \textit{natural language inference} in this way~\cite{schwarcz1970deductive,wilks1975preferential,punyakanok2004natural}.

The leading term \textit{recognizing} in RTE is fitting as the task is to classify or predict whether the truth of one sentence likely follows the other. The second term \textit{textual} is similarly appropriate since the domain is limited to textual data. %
Critics of the name RTE often argue that the term \textit{entailment} is inappropriate since the definition of the NLP task strays too far from the technical definition from \textit{entailment} in linguistics~\cite{manning2006local}.
\newcite{W05-1206} prefer the term \textit{textual inference} because examples in %
RTE datasets often require a system to not only identify entailments but also conventional implicatures, conversational implicatures, and world knowledge.

If starting over, we would advocate for the phrase \textit{Recognizing Textual Inference}. However, given the choice between RTE and NLI, we prefer RTE since it is more representative of the task at hand.

\section[Evaluating Natural Language Processing Systems]{Evaluating NLP Systems}
\label{sec:eval-background}
The question of how best to evaluate NLP systems is an open problem
intriguing the community for decades.
A 1988 workshop on the evaluation of NLP systems
explored key questions for evaluation. These included questions related to
valid measures of ``black-box'' performance, linguistic theories that
are relevant to developing test suites, %
reasonable expectations
for robustness, and %
measuring progress in the field~\cite{palmer-finin-1990-workshop}.
The large number of ACL workshops focused on evaluations in NLP demonstrate the lack
of consensus on how to properly evaluate NLP systems.
Some workshops focused on: 1) evaluations in general~\cite{ws-2003-eacl}; %
2) different NLP tasks, e.g. machine translation~\cite{ws-2001-mt,ws-2005-acl-intrinsic} and summarization~\cite{ws-2012-evaluation,ws-2017-multiling};
or 3) contemporary NLP approaches that rely on vector space representations~\cite{ws-2016-evaluating,ws-2017-evaluating,ws-2019-evaluating}.

In the quest to develop an ideal evaluation framework for NLP systems, researchers
proposed multiple evaluation methods, e.g.
EAGLES~\cite{eagles95}, TSNLP~\cite{Oepen95tsnlp,lehmann-etal-1996-tsnlp},
\fracas~\cite{cooper1996using},  SENSEVAL~\cite{kilgarriff1998senseval}, CLEF~\cite{agosti2007future}, %
and others.
These approaches are often divided along multiple dimensions.
Here, we will survey approaches along two dimensions: %
1) intrinsic vs. extrinsic evaluations;
2) general purpose vs task specific evaluations.\footnote{\newcite{resnik201011} summarize other evaluation approaches
and \newcite{paroubek2007principles} present a history and evolution of NLP evaluation methods.}

\subsection{Intrinsic vs Extrinsic Evaluations}
\label{subsec:intrinsic-extrinsic}
\begin{quote}
Intrinsic evaluations test the system in of itself and extrinsic evaluation test the 
system in relation to some other task.

\hfill~\cite{farzindar2004letsum}
\end{quote}

When reviewing \newcite{10.5555/547445}'s textbook on NLP evaluations, \newcite{estival1997karen}
comments that ``one of the most important distinctions that must be drawn when performing an
 evaluation of a system is that between \textit{intrinsic criteria}, i.e. those concerned with
 the system's own objectives, and \textit{extrinsic criteria}, i.e. those concerned with the
 function of the system in relation to its set-up.'' \newcite{resnik2006using} similarly noted that
``intrinsic evaluations measure the performance of an NLP component on its defined subtask, usually against a defined standard in a reproducible laboratory setting'' while ``extrinsic evaluations focus on the component’s contribution to the performance of a complete application, which often involves the participation of a human in the loop.''
\newcite{sparck-jones-1994-towards} refers to the distinction of intrinsic vs extrinsic evaluations as the \textit{orientation}
of an evaluation.

Under these definitions, for example, ``an intrinsic evaluation of a parser would
analyze the accuracy of the results returned by the
parser as a stand-alone system, whereas an extrinsic evaluation would analyze the impact of the
parser within the context of a broader NLP application'' like answer extraction~\cite{molla-hutchinson-2003-intrinsic}.
When evaluating a document summarization system, an intrinsic evaluation might ask
questions related to the fluency or coverage of key ideas in the summary while an 
extrinsic evaluation might explore whether a generated summary was useful in a search engine~\cite{resnik201011}.
This distinction has also been referred to as application-free versus 
application-driven evaluations~\cite{kovaz2016evaluation}.\footnote{
As another example,
in the case of evaluating different methods for training word
vectors, intrinsic evaluations
might consider how well similarities between word vectors correlate with human evaluated word 
similarities.
This is the basis of evaluation benchmarks like SimLex~\cite{hill-etal-2015-simlex}, Verb~\cite{baker-etal-2014-unsupervised}, RW~\cite{luong-etal-2013-better}, MEN~\cite{bruni2012distributional}, WordSim-353~\cite{finkelstein2001placing}, and others.
Extrinsic evaluations for word embeddings might consider how well different word vectors help models
for tasks like sentiment analysis~\cite{petrolito2018word,10.1007/978-3-030-33110-8_14}, machine translation~\cite{wang2019evaluating}, or  named entity recognition~\cite{wu2015study,nayak2016evaluating}.
}

Proper extrinsic evaluations are often infeasible in an academic lab setting.
Therefore, researchers
often rely on intrinsic evaluations to approximate extrinsic evaluations,
even though intrinsic and extrinsic evaluations serve different goals
and many common intrinsic evaluations
for word vectors~\cite{tsvetkov-etal-2015-evaluation,chiu-etal-2016-intrinsic,faruqui-etal-2016-problems}, generating natural language text~\cite{belz-gatt-2008-intrinsic,doi:10.1162}, or text mining~\cite{caporaso2008intrinsic} might not correlate with extrinsic evaluations.\footnote{Although recent work suggest that some intrinsic evaluations for
word vectors do indeed correlate with extrinsic evaluations~\cite{qiu2018revisiting,thawani-etal-2019-swow}.}
Developing intrinsic evaluations that correlate with extrinsic evaluations remains
an open problem in NLP.

\subsection{General Purpose vs Task Specific Evaluations}
General purpose evaluations determine how well NLP systems capture different linguistic phenomena.
These evaluations often rely on the development of test cases that systematically cover a wide range of phenomena. 
Additionally, these evaluations generally do not consider %
how well a system under investigation
performs on held out data for the task that the NLP system was trained on. 
In general purpose evaluations, specific linguistic phenomena should be
isolated such that each test or example evaluates one specific linguistic phenomenon,
as tests ideally %
``are controlled and exhaustive databases of linguistic utterances classified by linguistic 
features''~\cite{lloberes-etal-2015-suitability}. 

In task specific evaluations, the goal is to determine how well a model performs on a held out test corpus. 
How well systems generalize on text classification problems is determined with a combination of metrics like 
accuracy, precision, and recall, or metrics like BLEU~\cite{papineni-etal-2002-bleu} and ROUGE~\cite{lin-2004-rouge} in generation tasks.
Task specific evaluations, where
``the majority of benchmark datasets $\ldots$ are drawn from text corpora, reflecting a natural frequency distribution of language phenomena''~\cite{belinkov2019survey},
is the common paradigm in NLP research today. 
Researchers often begin their research with provided %
training and held-out test corpora,
as their research agenda is to develop systems %
that outperform other researchers' systems %
on a held-out test set %
based on a wide range of metrics. 

The distinction between general purpose and task specific evaluations is sometimes blurred.
For example, while general purpose evaluations are ideally task agnostic,
researchers develop %
evaluations that test for a wide range of linguistic phenomena captured by NLP systems trained to perform specific tasks. 
These include linguistic tests targeted for systems that focus on parsing~\cite{lloberes-etal-2015-suitability},
machine translation~\cite{king-falkedal-1990-using,koh2001test,isabelle-etal-2017-challenge,choshen-abend-2019-automatically,popovic-castilho-2019-challenge,avramidis-etal-2019-linguistic}, summarization~\cite{pitler-etal-2010-automatic}, and 
others~\cite{chinchor-1991-muc-3,chinchor-etal-1993-evaluating}.

\paragraph*{Test Suites vs. Test Corpora}
This distinction can also be described in terms of 
the data used to evaluate systems. \newcite{Oepen95tsnlp} 
refer to this distinction as test suites versus test corpora. They define
a test suite as a ``systematic collection of linguistic
expressions (test items, e.g. sentences or phrases) and
often includes associated annotations or descriptions.''
They lament the state of test suites in their time since  
``most of the existing test suites have been
written for specific systems or simply enumerate a set
of `interesting' examples%
[but] does not meet
the demand for large, systematic, well-documented
and annotated collections of linguistic material required by a growing number of NLP applications.'' 
\noindent
\citeauthor{Oepen95tsnlp} further delineate the difference between test corpora and test suites.
Unlike ``test corpora drawn from naturally occurring texts,'' test suites allow
for 1) more control over the data,
2) systematic coverage, 3) non-redundant representation, 4) inclusion of negative data, 
and 5) coherent annotation.
Thus, test suites ``allow for a fine-grained diagnosis of system performance''~\cite{Oepen95tsnlp}.
\citeauthor{Oepen95tsnlp} argue that both should be used
in tandem - ``test suites and corpora should stand in a complementary relation, with the former building on the latter wherever possible and necessary.'' Hence, 
both test suites and test corpora are important for evaluating how well NLP systems capture linguistic phenomena
and perform in practice on real world data.

\subsection{Probing Deep Learning NLP Models}
\label{sec:probes}
In recent years, interpreting and analysing NLP models has become prominent in many research agendas.
Contemporary and successful deep learning NLP methods are not as interpretable as previously popular NLP approaches relying on feature engineering. 
Approaches for interpreting and analysing how well NLP models capture linguistic
phenomena often leverage auxiliary or diagnostic classifiers.
Contemporary deep learning NLP systems
often leverage pre-trained encoders to represent the meaning of a sentence in a fixed-length
vector representation. \newcite{adi2017fine} introduced the notion of using auxiliary classifiers
as a general purpose methodology
to diagnose what language information is encoded and captured by contemporary sentence representations.
They argued for using ``auxiliary prediction tasks'' where, like in \newcite{dai2015semi}, pre-trained sentence 
encodings are ``used as input for other prediction tasks.'' 
The ``auxiliary prediction tasks'' can serve as diagnostics, and
\newcite{adi2017fine}'s auxiliary, diagnostic tasks focused 
on how word order, word content, and sentence length are captured in pre-trained sentence representations.

As \citeauthor{adi2017fine}'s general methodology ``can be applied to any sentence representation model,''
researchers %
develop other diagnostic tasks that explore different linguistic 
phenomenon~\cite{ettinger-etal-2018-assessing,conneau2018probe,hupkes2018visualisation}.
\newcite{belinkov2018internal}'s thesis relied on and popularized this methodology when exploring how well speech recognition
and machine translation systems capture phenomena related to phonetics~\cite{NIPS2017_6838}, 
morphology~\cite{P17-1080}, and syntax~\cite{belinkov-EtAl:2017:I17-1}.

The general purpose methodology of auxiliary diagnostic classifiers is also used to explore how well different pre-trained 
sentence representation methods perform on a broad range of NLP tasks. For example, 
SentEval~\cite{conneau-kiela-2018-senteval} and GLUE~\cite{wang2018glue} are used to evaluate how different sentence representations perform 
on paraphrase detection, semantic textual similarity, and a wide range of other binary and multi-class
classification problems. 
We categorize these datasets as extrinsic evaluations since they often
treat learned sentence-representations as features to train a classifier for an external task.
However, most of these do not count as test suites, since the data is not tightly controlled to evaluate specific
linguistic phenomena. Rather, resources like GLUE and SuperGLUE~\cite{wang2019superglue} package existing test corpora for different tasks and provide
an easy platform for researchers to compete on developing systems that perform well on the suite
of pre-existing, and re-packaged test corpora.

\section{Recognizing Textual Entailment}%

\begin{quote}
NLP systems cannot be held responsible for knowledge
of what goes on in the world but no NLP system can
claim to ``understand'' language if it can’t cope with
textual inferences.

\hfill ~\cite{W05-1206}
\end{quote}

Recognizing and coping with inferences is key to understanding %
human language.
While NLP systems might be trained to perform different tasks, such as translating, answering questions, or extracting
information from text, most NLP systems require understanding and making inferences from text.
Therefore, RTE was introduced as a framework to evaluate NLP systems.
Rooted in linguistics, RTE is the task of determining whether the meaning of one sentence can likely be inferred from another. 
Unlike the strict definition of entailment in linguistics that 
``sentence A entails sentence B if in all models in which the interpretation of A is true, also the interpretation of B is true''~\cite{sep-montague-semantics},
RTE relies on a fuzzier notion of entailment. For example, 
annotation guidelines for an RTE dataset\footnote{These were the guidelines in RTE-1.} stated that
\begin{quote}
in principle, the hypothesis must be fully
entailed by the text. Judgment would be
False if the hypothesis includes parts that
cannot be inferred from the text. However,
cases in which inference is very probable
(but not completely certain) are still judged
as True. 

\hfill~\cite{dagan2006pascal}
\end{quote}

\begin{table*}[t!]
\centering
\begin{tabular}{HHl|c}
\toprule
& \textbf{P} &	Kessler 's team conducted 60,643 interviews with adults in 14 countries \\
\multirow{-2}{*}{RTE1}& \textbf{H} & $\blacktriangleright$ Kessler 's team interviewed more than 60,000 adults in 14 countries & \multirow{-2}{*}{entailed}\\ 
\midrule 

& \textbf{P} &	Capital punishment is a catalyst for more crime \\
\multirow{-2}{*}{RTE2}& \textbf{H} & $\blacktriangleright$	Capital punishment is a deterrent to crime & \multirow{-2}{*}{not-entailed}\\ 
\midrule 
& &Boris Becker is a former professional tennis player for Germany & \\
\multirow{-2}{*}{RTE3} & & $\blacktriangleright$ Boris Becker is a Wimbledon champion & \multirow{-2}{*}{not-entailed}\\  
\bottomrule
\end{tabular}
\caption[Examples from PASCAL RTE datasets]{Examples from the PASCAL RTE datasets (modified for space): 
The first line in each example is the premise and the line starting with $\blacktriangleright$ is the corresponding hypothesis. The first, second, and third examples are from the RTE1, RTE2, and RTE3 development sets respectively.
The second column indicates the example's label.} %
\label{tab:pascal-example}
\end{table*}
\noindent
Starting with \fracas, we will
discuss influential work that introduced and argued for RTE as an evaluation framework.

\paragraph*{
\fracas}
Over a span of two years (December 1993 - January 1996),
\newcite{cooper1996using} developed \fracas as
``an inference test suite for evaluating the inferential competence
of different NLP systems and semantic theories''. 
Created manually
by many linguists and funded by FP3-LRE,\footnote{https://cordis.europa.eu/programme/id/FP3-LRE} \fracas is a ``semantic test suite'' that covers a range of semantic phenomena
categorized into 9
classes. These are 
generalized quantifiers, plurals, anaphora, ellipsis,
adjectives, comparatives, temporal reference, verbs, and attitudes. %
Based on the descriptions in ~\cref{sec:eval-background}, we would classify \fracas as an intrinsic evaluation and a general purpose test suite.

Examples in \fracas contain a premise paired with a hypothesis. Premises are at least one sentence,
though sometimes they contain multiple sentences, and most hypotheses
are written in the form of a question and the answers are either \textit{Yes}, \textit{No},
or \textit{Don't know}.
\newcite{maccartney2009natural} (specifically Chapter 7.8.1) converted the hypotheses from questions into declarative statements.\footnote{\url{https://nlp.stanford.edu/~wcmac/downloads/fracas.xml}}
\autoref{tab:fracas-example} (in the appendix) contains examples from \fracas.
In total, \fracas only contains about $350$ labeled examples, 
potentially limiting the ability to generalize how well models capture
these phenomena. Additionally, the limited number of examples in \fracas prevents 
its use as a dataset to train data hungry deep learning models.

\begin{table*}[t!]
\centering
\begin{tabular}{l|l|c}
\toprule
\textbf{P} &	\multicolumn{2}{c}{A woman is talking on the phone while standing next to a dog} \\
\textbf{H1} & 	A woman is on the phone & entailment \\
\textbf{H2} & 	A woman is walking her dog & neutral \\
\textbf{H3} & 	A woman is sleeping & contradiction \\ 
\midrule
\textbf{P} &	\multicolumn{2}{c}{Tax records show Waters earned around \$65,000 in 2000} \\
\textbf{H1} & 	 Waters' tax records show clearly that he earned a lovely \$65k in 2000 & entailment \\
\textbf{H2} & 	Tax records indicate Waters earned about \$65K in 2000& entailment \\ 
\textbf{H3} & 	Waters' tax records show he earned a blue ribbon last year & contradiction \\
\bottomrule
\end{tabular}
\caption[Examples from SNLI and Multi-NLI]{Examples from the development sets of SNLI (top) and MultiNLI (bottom).
Each example contains one premise that is paired with three hypotheses in the datasets.}
\label{tab:snli-example}
\end{table*}

\noindent
\paragraph*{Pascal RTE Challenges}

With a similar broad goal as \fracas, the Pascal Recognizing Textual Entailment challenges 
began as a ``generic evaluation framework” to compare the inference capabilities
of models designed to perform different tasks, based on the intuition ``that major inferences, as needed by
multiple applications, can indeed be cast in terms
of textual entailment''~\cite{dagan2006pascal}.
Unlike \fracas's goal of determining whether a model performs
distinct types of reasoning, the Pascal RTE Challenges primarily focused on using this 
framework to evaluate
models for distinct, real-world downstream tasks. Thus, the examples in the Pascal
RTE datasets were
extracted from downstream tasks. 
The process was referred to as \textit{recasting} in
the thesis by \newcite{glickman2006applied}.

NLU problems were reframed under the RTE framework and candidate sentence pairs were extracted
from existing NLP datasets and then labeled under variations of the RTE definition (including the quote above~\cite{dagan2006pascal}).\footnote{See ~\autoref{sec:pascal-definitions} for the annotation guidelines for RTE1, RTE2, and RTE3.}
For example, the RTE1 data came from $7$ tasks: comparable documents, reading comprehension,
question answering, information extraction, machine translation,
information retrieval, and paraphrase acquisition.\footnote{Chapter 3.2
of Glickman's thesis discusses how examples from these datasets were
converted into RTE.}
Starting with \newcite{dagan2006pascal}, there have been eight iterations of the %
RTE challenge,
with the most recent being \newcite{dzikovska-EtAl:2013:SemEval-2013}.

\paragraph*{SNLI and MNLI}
The most popular recent RTE datasets, Stanford Natural Language Inference~\cite[SNLI;][]{snli:emnlp2015}
and its successor Multi-NLI~\cite{williams2017broad}, 
each contain over half a million examples
and enabled researchers to apply data-hungry deep learning methods to RTE. 
Unlike the RTE datasets, these two datasets were created by eliciting hypotheses from 
humans.
Crowd-source workers were tasked with writing one sentence each that is entailed,
neutral, and contradicted by a
caption extracted from the Flickr30k corpus~\cite{young2014image}.
Next, the label for each premise-hypothesis pair in the development and test sets were verified by multiple crowd-source workers and the majority-vote label was assigned for each example.  
\autoref{tab:snli-example} provides such examples for both datasets. 
\newcite{rudinger-may-vandurme:2017:EthNLP} illustrated how eliciting textual data in this fashion creates stereotypical biases in SNLI.
Some of the biases are gender-, age-, and race-based. 
\newcite{Poliak2018StarSem} argue that this may cause additional biases enabling a hypothesis-only model to outperform the majority baseline on SNLI by 100 percent~\cite{Gururangan2018,Tsuchiya2018}.

\subsection{Entailment as a Downstream NLP Task}
The datasets in the PASCAL RTE Challenges were primarily treated as test corpora. Teams participated in those challenges by developing models to achieve increasingly high scores on each challenges' datasets. 
Since RTE was motivated as a diagnostic, researchers analyzed the RTE challenge datasets.
\newcite{P08-1118} argued that there exist different levels and types of contradictions. They focus on different types of
phenomena, e.g. antonyms, negation, and world knowledge, that can explain why a premise contradicts a hypothesis.
\newcite{maccartney2009natural} used a simple bag-of-words model to evaluate early iterations of Recognizing Textual Entailment (RTE) challenge sets
and noted\footnote{In Chapter 2.2 of his thesis} that ``the RTE1 test suite is the
hardest, while the RTE2 test suite is roughly 4\% easier, and
the RTE3 test suite is roughly 9\% easier.''
Additionally,
\newcite{vanderwende2006syntax} and \newcite{Blake:2007:RSS:1654536.1654557} demonstrate
how sentence structure alone can provide a high
signal for some RTE datasets.\footnote{\newcite{vanderwende2006syntax} explored RTE-1
and \newcite{Blake:2007:RSS:1654536.1654557} analyzed RTE-2 and RTE-3.}
Despite these analyses, researchers primarily built models to perform the task on the PASCAL RTE datasets rather than leveraging these datasets to evaluate models built for other tasks.

Coinciding with the recent ``deep learning wave'' that has taken over
NLP and Machine Learning~\cite{manning2015computational}, %
the introduction of large scale RTE datasets, specifically SNLI and MNLI, 
led to a resurgence of interest in RTE amongst NLP researchers. 
Large scale RTE datasets focusing on specific domains, like grade-school scientific 
knowledge~\cite{Khot2018} or medical information~\cite{romanov-shivade-2018-lessons}, emerged as well.
However, 
this resurgence did not primarily focus on using RTE as a means to evaluate NLP systems.
Rather, %
researchers primarily used these datasets to compete with one another to achieve the top score on leaderboards
for new RTE datasets.

\section{Revisiting RTE as an NLP Evaluation} %
\label{sec:rte-eval}
\begin{quote}
There has been little
evidence to suggest [that RTE models] capture the type of compositional or world knowledge tested by
datasets like the FraCas test suite.

\hfill~\cite{pavlick:2017:thesis}
\end{quote}

As large scale RTE datasets, like SNLI and MNLI, 
rapidly surged in popularity,
some researchers critiqued the datasets' ability to test the inferential capabilities
of NLP models. 
A high accuracy on these datasets does not
indicate which types of reasoning RTE models perform or capture.
As noted by \newcite{white-EtAl:2017:I17-1}, ``researchers compete on which system achieves the highest score on a test set, but this
itself does not lead to an understanding of which
linguistic properties are better captured by a quantitatively superior system.'' 
In other words, the single accuracy metric on these challenges indicates how well a model can recognize
whether one sentence likely follows from another, but it does not illuminate how well
NLP models capture different semantic phenomena that are important for general NLU.

This issue was pointed out regarding the earlier PASCAL RTE datasets.  In her thesis that
presented ``a test suite for adjectival inference developed as a resource for the evaluation
of computational systems handling natural language inference.''
\newcite{amoia:tel-01748535} blamed ``the difficulty of defining the linguistic phenomena which are responsible for inference'' as the reason why previous RTE resources ``concentrated on the creation of applications coping with textual entailment'' rather than ``resources for the
evaluation of such applications.''

As current studies began exploring what linguistic phenomena are captured by neural NLP models and auxiliary diagnostic classifiers became a common tool to evaluate sentence representations in NLP systems, (\cref{sec:probes}), the community saw a interest in developing RTE datasets that can provide insight into what type of linguistic phenomena are captured by neural, deep learning models.
In turn, the community is answering \newcite{W17-7203}  plea to the community to test ``more kinds of inference'' than in previous RTE
challenge sets.
Here, we will highlight recent efforts in creating datasets that demonstrate how the community has started answering \citeauthor{W17-7203}'s call.
We group these different datasets based on how they were created.
and \autoref{tab:diagnostic-rte-datasets} includes additional RTE datasets focused on specific linguistic phenomena.

\subsection{Automatically Created}
\newcite{white-EtAl:2017:I17-1} advocate for using RTE as a single framework
to evaluate different linguistic phenomena. They argue for creating RTE datasets focused on specific phenomena by \textit{recasting} existing annotations for different semantic phenomena into RTE. \newcite{D18-1007} introduce the Diverse Natural Language Inference Collection (DNC) of over half a million RTE examples. They create the DNC by converting 7 semantic phenomena from 13 existing datasets into RTE. These phenomena include event factuality, named entity recognition, gendered anaphora
resolution, sentiment analysis, relationship extraction, pun detection, and lexicosyntactic inference. 
\newcite{staliunaite2018learning}'s master's thesis improved \newcite{D18-1007}'s method used to recast annotations for factuality into RTE.
Other efforts have created recast datasets in Hindi that focus on sentiment and emotion detection.\footnote{\url{https://github.com/midas-research/hindi-nli-data}}

Concurrent to the DNC, \newcite{naik-EtAl:2018:C18-1} released the ``NLI Stress Tests'' that included RTE datasets focused on negation, word overlap between premises and hypotheses, numerical reasoning, amongst other phenomena. \newcite{naik-EtAl:2018:C18-1} similarly create their stress tests automatically using different methods for each phenomena. They then used these datasets to evaluate how well a wide class of RTE models capture these phenomena.   
Other RTE datasets that target more specific phenomena %
were created using automatic methods, including \newcite{jeretic2020natural}'s ``IMPRES'' diagnostic RTE dataset that tests for IMPlicatures and PRESuppositions.

If not done with thorough testing and care, recasting or other automatic methods for creating these RTE datasets can lead to annotation artifacts unrelated to RTE that limit how well a dataset tests for a specific semantic phenomena. For example, to create not-entailed hypotheses, \newcite{white-EtAl:2017:I17-1} replaced a single token in a context sentence with a word that crowd-source workers labeled as not being a paraphrase of the token in the given context. In FN+~\cite{pavlick-EtAl:2015:ACL-IJCNLP2}, two words might be deemed to be incorrect paraphrases in context based on a difference in the words' part of speech tags.\footnote{ 
\autoref{tab:bad-paraphrases-examples} (in the appendix) demonstrates such examples, and in the last example, the words ``on'' and ``dated'' in the premise and hypothesis respectively have the \texttt{NN} and \texttt{VBN} POS tag.} This limits the utility of the recast version of FN+ to be used when evaluating how well models capture paraphrastic inference.

Similar to the efforts described here to recast different NLU problems as RTE,
others have recast NLU problems into a question answer format~\cite{mccann2018natural,gardner2019question}.
Recasting problems into RTE, as opposed to question-answering, has deeper roots
in linguistic theory~\cite{westernling1998,chierchia2000meaning,Brinton2000TheSO}, and continues a rich history within the NLP community.

\subsection{Semi-Automatically Created}
Other RTE datasets focused on specific phenomena rely on semi-automatic methods. RTE pairs are often generated automatically using well developed heuristics. Instead of automatically labeling the RTE example pairs (like in the approaches previously discussed), the automatically created examples are often labeled by crowdsource workers. For example, \newcite{kim-etal-2019-probing} use hueristics to create RTE pairs that test for prepositions, comparatives, quantification, spacial reasoning, and negation and then present these examples to crowdsource workers on Amazon Mechanical Turk.
Similarly, 
\newcite{ross-pavlick-2019-well} generate two premise-hypothesis pairs for each RTE example in MNLI that satisfy their set of constraints. Next, they rely on crowdsource workers to annotated whether the premise likely entails the hypothesis on a $5$-point Likert scale. 

Some methods instead first manually annotate their data and then rely on automatic methods to construct hypotheses and label RTE pairs. 
When generating RTE examples testing for monotonicity, \newcite{Richardson2020ProbingNL} first manually encode the ``monotonicity information of each token in the lexicon
and built sentences via a controlled set of grammar rules.'' They then ``substitute upward entailing tokens or
constituents with something `greater than or equal to' 
them, or downward entailing ones with something `less than
or equal to' them." 

\begin{table*}[t!]
    \centering
    \begin{tabular}{p{16cm}HH}
    \toprule
        Proto-Roles~\cite{white-EtAl:2017:I17-1}, Paraphrastic Inference~\cite{white-EtAl:2017:I17-1},
        Event Factuality~\cite{D18-1007,staliunaite2018learning}, Anaphora Resolution~\cite{white-EtAl:2017:I17-1,D18-1007},
        Lexicosyntactic Inference~\cite{P16-1204,D18-1007,glockner-shwartz-goldberg:2018:Short}, Compositionality~\cite{dasgupta2018evaluating},
        Prepositions~\cite{kim-etal-2019-probing},
        Comparatives~\cite{kim-etal-2019-probing,Richardson2020ProbingNL},
        Quantification/Numerical Reasoning~\cite{naik-EtAl:2018:C18-1,kim-etal-2019-probing,Richardson2020ProbingNL}, 
        Spatial Expressions~\cite{kim-etal-2019-probing},
        Negation~\cite{naik-EtAl:2018:C18-1,kim-etal-2019-probing,Richardson2020ProbingNL},
        Tense \& Aspect~\cite{kober-etal-2019-temporal},
        Veridicality~\cite{D18-1007,ross-pavlick-2019-well},
        Monotonicity~\cite{yanaka-etal-2019-neural, yanaka2020neural,Richardson2020ProbingNL}, 
        Presupposition~\cite{jeretic2020natural}, 
        Implicatures~\cite{jeretic2020natural}, Temporal Reasoning~\cite{Vashishtha:emnlp:2020} \\
        \bottomrule
    \end{tabular}
    \caption{List of different semantic phenomena tested for in recent RTE datasets.} %
    \label{tab:diagnostic-rte-datasets}
\end{table*}

\subsection{Manually Created}
While most of these datasets rely on varying degrees of automation, some RTE datasets focused on evaluating how well models capture specific phenomena rely on manual annotations. The GLUE and SuperGlue datasets include diagnostic sets where annotators manually labeled samples of examples as requiring a broad range of linguistic phenomena. The types of phenomena manually labeled include lexical semantics, predicate-argument structure, logic, and common sense or world knowledge.\footnote{\url{https://gluebenchmark.com/diagnostics}}

\section{Recommendations}
These efforts %
resulted in a
consistent format and framework for testing how well contemporary, deep learning NLP systems
capture a wide-range of linguistic phenomena.
However, so far, most of these datasets that target specific linguistic phenomena have been used to solely evaluate how well RTE models capture a wide range of phenomena, as opposed to evaluating how well systems trained for more applied NLP tasks capture these phenomena. Since RTE was introduced as a framework to evaluate how well NLP models cope with inferences, these newly created datasets have not been used to their full potential.

A limited number of studies used some of these datasets to evaluate how well models trained for other tasks capture these phenomena. \newcite{poliakNAACL18} evaluated how well a BiLSTM encoder trained as part of a neural machine translation system capture phenomena like semantic proto-roles, paraphrastic inference, and anaphora resolution. \newcite{kim-etal-2019-probing} used their RTE datasets focused on function words to evaluate different encoders trained for tasks like CCG parsing, image-caption matching, predicting discourse markers, and others.  Those studies relied on the use of auxiliary classifiers as a common probing technique to evaluate sentence representations. 
As the community's interest in analyzing deep learning systems increases, demonstrated by the recent work relying on~\cite{ws-2018-2018,ws-2019-2019-acl} and improving upon~\cite{hewitt-liang-2019-designing,voita2020informationtheoretic,pimentel-etal-2020-information,mu2020compositional} the popular auxiliary classifier-based diagnostic technique,
we call on the community to leverage the increasing number of RTE datasets focused on different semantic phenomena (\autoref{tab:diagnostic-rte-datasets}) to thoroughly study the representations learned by downstream, applied NLP systems.
The increasing number of RTE datasets focused on different phenomena can help researchers use one standard format to analyze how well models capture different phenomena. 

Another recent line of work uses RTE to evaluate the output of text generation systems. For example, \newcite{falke-etal-2019-ranking} explore ``whether textual entailment predictions can be used to detect errors'' in abstractive summarization systems and if errors ``can be reduced by reranking alternative predicted summaries'' with a textual entailment system trained on SNLI.
While \newcite{falke-etal-2019-ranking} results demonstrated that current models might not be accurate enough to rank generated summaries, \newcite{barrantes2020adversarial} demonstrate that contemporary transformer models trained on the Adversarial NLI dataset~\cite{nie-etal-2020-adversarial} ``achieve significantly higher accuracy and have the potential
of selecting a coherent summary.'' Therefore, we are encouraged that researchers might be able to use many of these new RTE datasets focused on specific phenomena to evaluate the coherency of machine generated text based on multiple linguistic phenomena that are integral to entailment and NLU. 
This approach can help researchers use the RTE datasets to evaluate a wider class of models, specifically non-neural models, unlike the auxiliary classifier or probing methods previously discussed. 

The overwhelming majority, if not all, of these RTE datasets targeting specific phenomena rely on categorical RTE labels, following the common format of the task. However, 
as \newcite{chen2020uncertain} recently illustrated, categorical RTE labels do not capture the subjective nature of the task. Instead, they argue for scalar RTE labels that indicate how likely a hypothesis could be inferred by a premise. 
\newcite{pavlick-kwiatkowski-2019-inherent} similarly lament how labels are currently used in RTE datasets. \citeauthor{pavlick-kwiatkowski-2019-inherent} demonstrate that 
 a single label aggregated from multiple annotations for one RTE example minimizes the ``type of uncertainty present in [valid] human
disagreements.'' Instead, they argue that a
``representation should be evaluated in
terms of its ability to predict the full distribution
of human inferences (e.g., by reporting crossentropy against a distribution of human ratings),
rather than to predict a single aggregate score
(e.g., by reporting accuracy against a discrete majority label or correlation with a mean score).''
Future RTE datasets targeting specific phenomena that contain scalar RTE labels from multiple annotators (following \newcite{chen2020counterfactual}'s and \newcite{pavlick-kwiatkowski-2019-inherent}'s recommendations) can provide even more insight into contemporary NLP models.

\section{Conclusion}
With the current zeitgeist of NLP research where researchers are interested in analyzing state-of-the-art deep learning models, %
now is a prime time to revisit RTE as a method to
evaluate the inference capabilities of NLP models.
In this survey, we discussed recent advances in RTE datasets that focus on specific linguistic phenomena that are integral for determining whether one sentence is likely inferred by another. Since RTE was primarily motivated as an evaluation framework, we began this survey with a broad overview of prior approaches for evaluating NLP systems. This included the distinctions between instrinsic vs extrinsic evaluations and general purpose vs task specific evaluations.

We discussed foundational RTE datasets that greatly impacted the NLP community and included critiques of why they do not fulfill the promise of RTE as an evaluation framework.
We highlighted recent efforts to create RTE datasets that focus on specific linguistic phenomena. 
By using these datasets to evaluate sentence representations from neural models or rank generated text from NLP systems,  
researchers can help fulfil the promise of RTE as unified evaluation framework.
Ultimately, this will help us determine how well models understand language on a fine-grained level. %

\section*{Acknowledgements}
The author would like to thank the anonymous reviewers for their very helpful comments, 
Benjamin Van Durme, Aaron Steven White, and Jo\~ao Sedoc for discussions that shaped this survey, Patrick Xia and 
Elias Stengel-Eskin for feedback on this draft, and Yonatan Belinkov and Sasha Rush for the encouragement to write a survey on RTE. %

\bibliographystyle{acl_natbib}
\bibliography{emnlp2020,thesis}

\newpage
\appendix

\section{Pascal RTE Annotation Guidelines}
\label{sec:pascal-definitions}
In the first iteration of the PASCAL RTE challeges, the task organizers were frank in their view that they expected the task definition to change over time. They wrote that  ``finally, the task definition and evaluation
methodologies are clearly not mature yet. We expect them to change over time and hope that participants' contributions, observations and
comments will help shaping this evolving research
direction.''
Here, we include snippets from the annotation guidelines for the first three PASCAL RTE challenges:

\subsection{RTE1 Guidelines}
\textit{ Given that the text and hypothesis might
originate from documents at different
points in time, tense aspects are ignored.
 In principle, the hypothesis must be fully
entailed by the text. Judgment would be
False if the hypothesis includes parts that
cannot be inferred from the text. However,
cases in which inference is very probable
(but not completely certain) are still judged
at True. 
$\ldots$ To reduce the risk of unclear cases,
annotators were guided to avoid vague examples for which inference has some positive probability that is not clearly very
high.
 To keep the contexts in T and H self contained annotators replaced anaphors
with the appropriate reference from preceding sentences where applicable. They
also often shortened the hypotheses, and
sometimes the texts, to reduce complexity.}

\hfill \cite{dagan2006pascal}

\subsection{RTE2 Guidelines} 
\textit{The data collection and
annotation guidelines were revised and expanded $\ldots$
We say that t entails h if, typically, a
human reading t would infer that h is most likely
true. This somewhat informal definition is based
on (and assumes) common human understanding of
language as well as common background knowledge. Textual entailment recognition is the task of
deciding, given t and h, whether t entails h.
Some additional judgment criteria and guidelines
are listed below: %
\begin{itemize}
    \item  Entailment is a directional relation. The hypothesis must be entailed from the given text,
but the text need not be entailed from the hypothesis.
\item The hypothesis must be fully entailed by the
text. Judgment would be NO if the hypothesis
includes parts that cannot be inferred from the
text.
\item Cases in which inference is very probable (but
not completely certain) are judged as YES. For
instance, in pair \#387 one could claim that although Shapiro’s office is in Century City, he
actually never arrives to his office, and works
elsewhere. However, this interpretation of t is
very unlikely, and so the entailment holds with
high probability. On the other hand, annotators were guided to avoid vague examples for
which inference has some positive probability
which is not clearly very high.
\item Our definition of entailment allows presupposition of common knowledge, such as: a com-
pany has a CEO, a CEO is an employee of the
company, an employee is a person, etc. For instance, in pair \#294, the entailment depends on
knowing that the president of a country is also
a citizen of that country.
\end{itemize}
}

\hfill \cite{BarHaim2006TheSP}

\subsection{RTE3 Guidelines}
\textit{As entailment is a directional relation, the
hypothesis must be entailed by the given
text, but the text need not be entailed by
the hypothesis.
\begin{itemize}
\item The hypothesis must be fully entailed by
the text. Judgment must be NO if the hypothesis includes parts that cannot be inferred from the text.
\item Cases in which inference is very probable
(but not completely certain) were judged as
YES.
\item Common world knowledge was assumed,
e.g. the capital of a country is situated in
that country, the prime minister of a state is
also a citizen of that state, and so on. 
\end{itemize}
} %

\hfill \cite{giampiccolo-etal-2007-third}

\begin{table*}[t!]
\begin{adjustbox}{width=1\linewidth}%
\centering
\begin{tabular}{l|l}
\toprule
\multicolumn{2}{c}{QUANTIFIERS (14)} \\
\textbf{P} &	Neither leading tenor comes cheap. One of the leading tenors is Pavarotti. \\
\textbf{Q} &	Is Pavarotti a leading tenor who comes cheap? \\
\textbf{H} &	Pavarotti is a leading tenor who comes cheap. \\
\textbf{A} & No \\
\midrule
\multicolumn{2}{c}{PLURALS (94)} \\
\textbf{P} &	The inhabitants of Cambridge voted for a Labour MP. \\
\textbf{Q} & 	Did every inhabitant of Cambridge vote for a Labour MP? \\
\textbf{H} & 	Every inhabitant of Cambridge voted for a Labour MP. \\
\textbf{A} & 	Unknown \\ \midrule
\multicolumn{2}{c}{COMPARATIVES (243)} \\
\textbf{P} & ITEL sold 3000 more computers than APCOM. APCOM sold exactly 2500 computers. \\
\textbf{Q} & Did ITEL sell 5500 computers? \\
\textbf{H} & ITEL sold 5500 computers. \\
\textbf{A} & Yes \\ 
\bottomrule
\end{tabular}
\end{adjustbox}
\caption[Examples from Fracas]{Examples from Fracas: \textbf{P} represents the premise(s), \textbf{Q} represents
the question from \fracas, \textbf{H} represents the declarative statement \newcite{maccartney2009natural} created
and, \textbf{A} represents the label. The number in the parenthesis indicates the example ID from \fracas.}
\label{tab:fracas-example}
\end{table*}

\begin{table*}[t!]
\centering
\begin{tabular}{l}
\toprule
unemployment is at an all-time \underline{low}  \\
$\blacktriangleright$ unemployment is at an all-time \underline{poor} \\ \midrule
aeoi 's activities and \underline{facility} have been tied to several universities \\
$\blacktriangleright$ aeoi 's activities and \underline{local} have been tied to several universities \\ \midrule
jerusalem fell to the ottomans in 1517 , remaining under their \underline{control} for 400 years \\
$\blacktriangleright$ jerusalem fell to the ottomans in 1517 , remaining under their \underline{regulate} for 400 years \\ \midrule
usually such parking spots are \underline{on} the side of the lot \\
$\blacktriangleright$ usually such parking spots are \underline{dated} the side of the lot \\ %
\bottomrule
\end{tabular}
\caption[Ungrammatical or disfluent recast FN+ examples]{Not-entailed examples from FN+'s dev set where the hypotheses are ungrammatical. The first line in each section is a premise and the
lines with $\blacktriangleright$ are corresponding hypotheses. \underline{Underline words} represent
the swapped paraphrases.}
\label{tab:bad-paraphrases-examples}
\end{table*}

\end{document}